%% file: root.tex
\documentclass[letterpaper, 10 pt, conference]{ieeeconf}  % Comment this line out if you need a4paper

\usepackage{multicol}
\usepackage{commath}
\usepackage{amssymb}
\usepackage{siunitx}
\usepackage{graphicx}
\usepackage{xcolor}
\usepackage{amsmath}
\usepackage{hyperref}

\IEEEoverridecommandlockouts                              % This command is only needed if 
\title{\LARGE \bf
Toolbox Release: A WiFi-Based Relative Bearing Sensor for Robotics
}

\author{Ninad Jadhav$^{1}$, Weiying Wang$^{1}$, Diana Zhang$^{2}$, Swarun Kumar $^{2}$ and Stephanie Gil$^{1}$

\thanks {$^{1}$REACT Lab, Harvard, \{njadhav,weiyingwang,sgil\}@g.harvard.edu}%
\thanks{$^{2}$WiTech Lab, CMU,dianaz1@andrew.cmu.edu,swarun@cmu.edu}%
\thanks{This work was funded by Lincoln Labs Line grant, Sloan Research Fellowship 2021 (FG-2020-13998) and NSF awards (grant numbers: 1845225, 1718435 and 1837607).}
}

\begin{document}

\maketitle
\thispagestyle{empty}
\pagestyle{empty}

% \input{text/1_intro_v6.tex} %SG edit
% \input{text/1_intro_v7_cutting.tex} %WW edit
\input{text/1_intro.tex}

% \input{text/2_related_works.tex}

\input{text/3_SAR_primer.tex}

\input{text/4_system_architecture.tex}

\input{text/5_results.tex}

% \input{text/6_dataset.tex}

\input{text/7_conclusion.tex}


\bibliographystyle{IEEEtran}
\bibliography{IEEEabrv,references}

\end{document}

%% file: text/1_intro.tex
%%%%%%%%%%%%%%%%%%%%%%%%%%%%%%%%%%%%%%%%%%%%%%%%%%%%%%%%%%%%%%%%%%%%%%%%%%%%%%%%
\begin{abstract}
This paper presents the WiFi-Sensor-for-Robotics (WSR) toolbox, an open source C++ framework. It enables robots in a team to obtain relative bearing to each other, even in non-line-of-sight (NLOS) settings which is a very challenging problem in robotics. It does so by analyzing the phase of their communicated WiFi signals as the robots traverse the environment. This capability, based on the theory developed in our prior works, is made available for the first time as an opensource tool. It is motivated by the lack of easily deployable solutions that use robots' local resources (e.g WiFi) for sensing in NLOS. This has implications for localization, ad-hoc robot networks, and security in multi-robot teams, amongst others. The toolbox is designed for distributed and online deployment on robot platforms using commodity hardware and on-board sensors. We also release datasets demonstrating its performance in NLOS and line-of-sight (LOS) settings for a multi-robot localization usecase. Empirical results show that the bearing estimation from our toolbox achieves mean accuracy of 5.10 degrees. This leads to a median error of 0.5m and 0.9m for localization in LOS and NLOS settings respectively, in a hardware deployment in an indoor office environment. 
% \vspace{-0.05in}
\section*{Supplementary Material}
% \vspace{-0.05in}
\centering Code: \href{https://github.com/Harvard-REACT/WSR-Toolbox}{https://github.com/Harvard-REACT/WSR-Toolbox} \\
Demo Videos: \href{https://git.io/JuKOS}{https://git.io/JuKOS}
\end{abstract}

%%%%%%%%%%%%%%%%%%%%%%%%%%%%%%%%%%%%%%%%%%%%%%%%%%%%%%%%%%%%%%%%%%%%%%%%%%%%%%%%
% \vspace{-0.05in}
\section{Introduction}
% \vspace{-0.05in}
Estimating and/or sensing relative bearing between robots is important for many multirobot tasks such as coverage, rendezvous, and distributed mapping amongst others~\cite{RobustSLAM_Luca,multiagentExploration_BurgardThrun,persistentSurveillance,Chen2015AdaptiveLF,cooperativeMapping_leonard}. This typically requires using external infrastructure such as GPS, a pre-deployed infrastructure such as wireless tags or beacons, or knowledge of a shared map for localization~\cite{Prorok2014AccurateIL}. Additionally, robots often need to rely on their \emph{local} sensors such as cameras and LiDAR for sensing~\cite{Martinelli2020CooperativeVO}. As such, when operating in unknown or GPS-denied environments with walls and other occlusions i.e non-line-of-sight (NLOS), obtaining relative bearing is very challenging. Motivated by this problem, we create and release the WiFi-Sensor-for-Robotics (WSR) Toolbox that uses robots' on-board resources. It analyzes the WiFi signals transmitted between robots at $\approx$ 5 kB/sec, similar to lightweight ping packets, to obtain their relative bearing. 

%====AOA Profile Idea====
\begin{figure}[h]
  \centering
  \includegraphics[width=8.5cm,height=5.5cm]{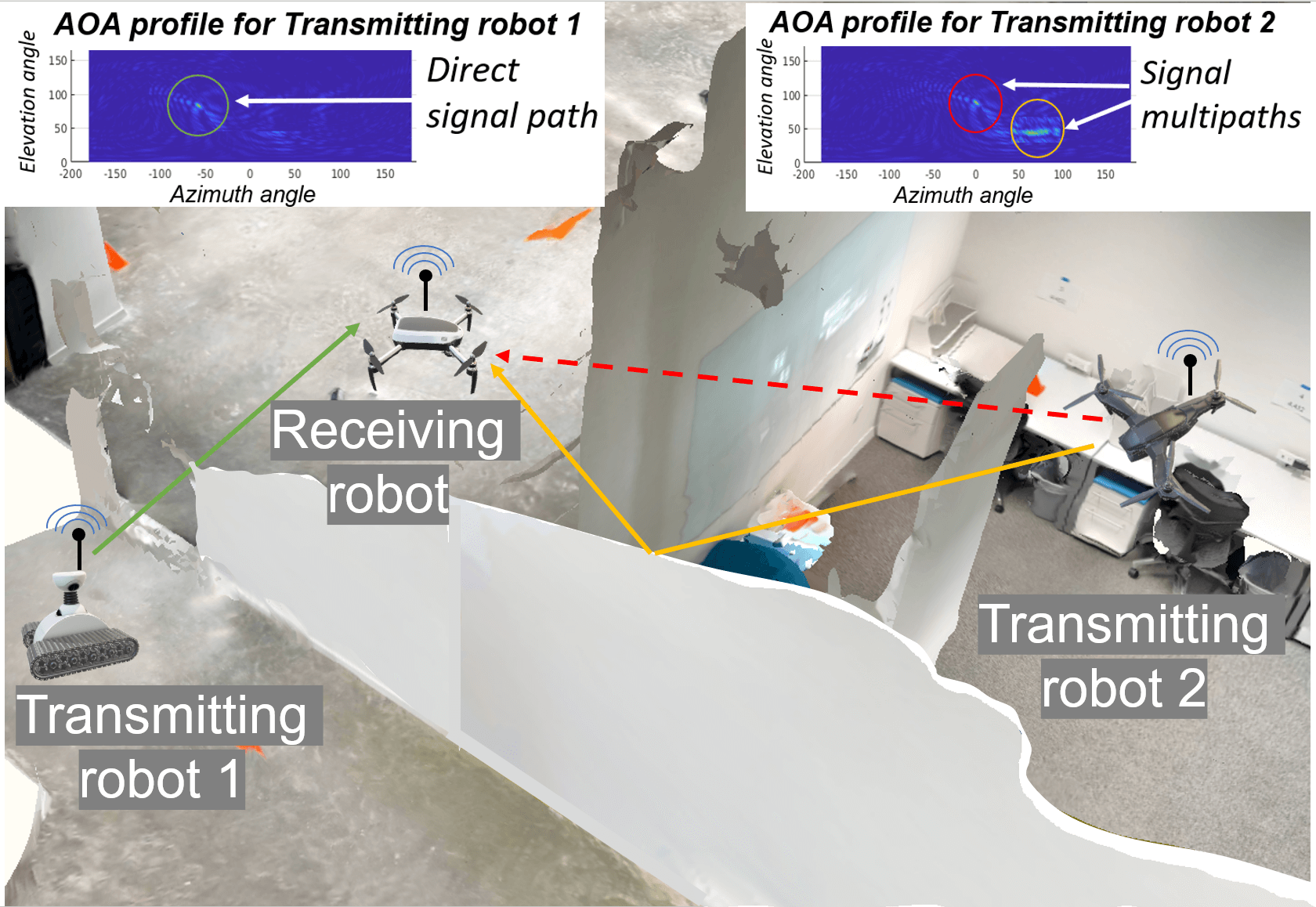}
  \vspace{-0.1in}
  \caption{\footnotesize{\emph{Schematic shows the multipath for WiFi signals transmitted between robots. Green, red and orange lines represent the line-of-sight, direct and reflected signal paths respectively. These multipaths are measured in the corresponding  Angle-Of-Arrival profiles as distinct peaks (highlighted circles). Our toolbox allows robots to obtain relative bearing i.e Angle-of-Arrival, using the highest magnitude signal peak in the profile shown below.}}}
  \label{fig:AOA_Profile}
  \vspace{-0.15in}
\end{figure}
%====AOA Profile Idea====
%====================AOA_profile==============
\begin{figure}
    \centering
    \begin{minipage}{0.75\linewidth}
     \includegraphics[width=7.0cm,height=3.0cm]{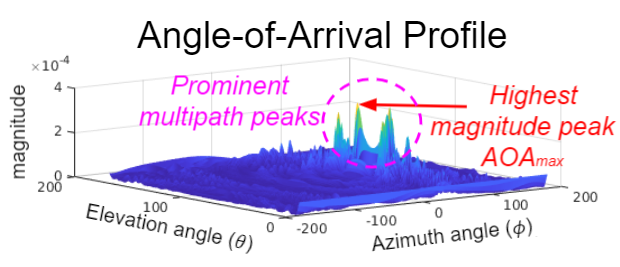}
    \end{minipage}
      \vspace{-0.1in}
    \caption{\footnotesize{
    \emph{Shows the outputs of the toolbox -- namely AOA profile, $AOA_{max}$ i.e relative bearing between robots in x-y (azimuth) and x-z plane (elevation) and prominent signal multipaths. Angles are in degrees.}}}
    \label{fig:real_aoa_profile}
    \vspace{-0.3in}
\end{figure}
%====================AOA-profile==============

WiFi has been studied as a sensing mechanism in NLOS for many applications that exploit its ability to traverse walls and occlusions. These applications include tracking~\cite{fadel3dTracking,fadelSeeThroughWalls}, imaging~\cite{mostofi_throughWall_wirelessSensing} and localization~\cite{ubicarse,Vasisht2016DecimeterLevelLW} amongst others. A wireless signal transmitted  between robots is attenuated, scattered, and reflected over several paths as it travels through the environment in a phenomenon called \emph{multipath} (Fig.~\ref{fig:AOA_Profile}). We refer to the measurement of these multipaths as an \emph{Angle-of-Arrival (AOA) profile} (Fig. \ref{fig:real_aoa_profile}). This AOA profile reveals information about the robots such as their relative bearing to one another, or their uniqueness, with implications for localization~\cite{ubicarse}, rendezvous and mapping~\cite{Wang2019ActiveRF}, maintaining communication quality in robot networks~\cite{Gil2015AdaptiveCI}, and security using AOA profile fingerprints~\cite{Gil2015Spoof-Resilient} amongst others.

Our previous works develop a method akin to Synthetic Aperture Radar (SAR) to obtain the AOA profile~\cite{Gil2015AdaptiveCI,ubicarse}. Essentially, this method requires using a robots' local displacement to take multiple measurements of the WiFi signals transmitted by the neighboring robot, to compute the profile. Our most recent work in~\cite{WSR_IJRR} solves key algorithmic challenges for making this method compatible with general robotic platforms including: i) implementing a SAR-like approach for robot displacements along an arbitrary trajectory in $\mathbb{R}^3$ in contrast to linear or turn-in-place circular motion as in prior works~\cite{Gil2015AdaptiveCI, Wang2019ActiveRF}, and ii) allowing for compatibility with noisy on-body inertial sensors (e.g. visual odometer), for estimating local displacement. However, none of our prior work made these capabilities available in a toolbox form for general accessibility of the robotics community.

This paper makes SAR-based relative bearing estimation using an AOA profile from~\cite{WSR_IJRR} available to the robotics community for the first time. Additional toolbox features that make it amenable to the community include a) C++ framework with a Cython wrapper to extend toolbox functionalities to python based applications, and b) an empirical characterization of the performance tradeoffs for online computation on on-board computers such as the UP-squared board. We hope to enable a new suite of perception capabilities based on the WSR toolbox and thereby enable the community to build upon this functionality with minimal deployment overhead. 
%====================AOA_profile==============
\begin{figure}
    \centering
    \begin{minipage}{0.95\linewidth}
        \includegraphics[width=8.0cm,height=3.4cm]{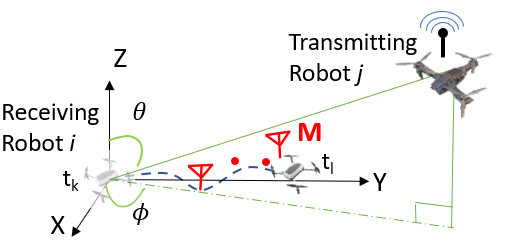}
    \end{minipage}
    % \vspace{-0.075in}
    \caption{\footnotesize{
    \emph{Robot $i$ collects $M$ WiFi packets along its path in $\mathbb{R}^3$ from time $t_k$ to $t_l$. Robot $i$ emulates a virtual antenna array using its motion and measures relative bearing in azimuth ($\phi$) and elevation ($\theta$) to robot $j$.}}}
    \label{fig:coordinate_frame}
    \vspace{-0.3in}
\end{figure}
%====================AOA-profile==============

We also demonstrate the applicability of the WSR toolbox for relative localization in multi-robot systems in NLOS which is a notoriously difficult problem in robotics. For our localization application, the transmitting ground robots are dispersed across six rooms in a real indoor office environment. Here, a localizing ground robot measures its relative location to several transmitting robots that are in NLOS, without a shared map or known visual landmarks. Using all on-board sensors, the AOA accuracy in the x-y plane achieved by our toolbox for these experiments is 5.10 degrees with a standard deviation of 25.16 degrees. This results in median localization error of 0.9m for $\approx$ 8m inter-robot distances in NLOS. To further enable follow-on work, we also release the datasets of our localization experiments as part of this toolbox. These include robot displacements and WiFi data for localization in NLOS as well as LOS. The toolbox also enables other applications like blind rendezvous between heterogeneous robots (see supplementary material).

\noindent \textbf{Related Work.} 
Extensive work has been done in the wireless community for localizing communicating devices \cite{Kotaru2015SpotFiDL,Kumar2014AccurateIL,securearray,spotFi,Xiong2013ArrayTrackAF}. Going beyond positional information, wireless signals have been used for material sensing \cite{swarun_materialSensing}, tracking humans through the wall \cite{Adib2015ThroughWall} and wireless imaging \cite{Guan2020ThroughFH} among others. However deploying these approaches on mobile robot platforms is challenging due to constraints on size, weight and power (SWaP). Recent works are exploring ways to utilize wireless signal information more efficiently in robotics. A Radio-Frequency Identification (RFID) based system for multi-robot scenarios is introduced in \cite{Zhou2009RFIDLA}, but requires predeployed infrastructure (RFID nodes). Ultra-Wide Band (UWB) have been used for localization and mapping either by using custom radio modules as anchors or by fusing with RGBD camera data \cite{uwb_localization,uwb_rgbd_sensing}. However, these sensors are not native to today's robot platforms. They typically need to operate on wide bandwidths resulting in significant FCC restrictions on operating range and power~\cite{fccban_uwb}. 

WiFi and bluetooth are more ubiquitous on today's robot platforms; methods utilizing their signal strength (RSSI) are more common in robotics~\cite{Coppola2018OnboardCR,Amoolya2019WiFiRB}. However RSSI requires sampling along multiple directions with substantial robot displacement to capture change and are coarse due to the impact of noise or deep fades \cite{Yan2013CoOptimizationOC,Zafari2019ASO}. Using signal phase to obtain bearing shows promise beyond using RSSI alone. The phase of a 5GHz WiFi changes by 2$\pi$ for every 6cm of robot displacement, far exceeding any change that might be attributed to noise variability. Thus, phase measurements enable higher granularity and accuracy in estimation of spatial information \cite{Zafari2019ASO}. Previous works that use phase however, constrain robot motion~\cite{Gil2015AdaptiveCI,Wang2019ActiveRF}. On the other hand, commercial blackbox bluetooth sensors that also use signal phase provide only bearing information. 

Our toolbox uses a WiFi signal's phase and offers more flexibility in terms of usage and deployment on multiple robots. It uses low data transmission rates which is useful when the robots are separated by long distances or when communication is constrained. Being opensource, a user of our toolbox has complete access to all the functionalities associated with bearing estimation. The AOA profile obtained from our toolbox can additionally be utilized for detecting outlying measurements~\cite{Wang2019ActiveRF} and applications beyond bearing estimation like adhoc robot networks, improving resilience and security in robot teams  \cite{Gil2015AdaptiveCI,Gil2015Spoof-Resilient,crowd_vetting,securearray}.  

\smallskip
\noindent \textbf{Contributions:}  This paper contributes the following:
\begin{itemize}
    \item  We release the WSR toolbox a C++ framework with Cython wrapper that allows robots to obtain relative bearing to each other. This toolbox processes data from a robots' on-body sensors (local displacement and received WiFi packets) on its on-board computer.
    \item We analyze the toolbox's performance and showcase its utility using a multi-robot localization scenario for both LOS and NLOS settings in an indoor office building.
    \item We provide experimental evaluation of AOA accuracy and release datasets of our hardware experiments.  
\end{itemize}

%% file: text/3_SAR_primer.tex
% \vspace{-0.075in}
\section{SAR Primer} \label{sec:primer_sar}
% \vspace{-0.05in}
We briefly describe the SAR method implemented in the toolbox and point the reader to~\cite{WSR_IJRR} for more details. SAR takes advantage of a robots' displacement and the WiFi signals it receives, to estimate AOA. Given a team of robots, we denote any robot that is receiving the signal as the robot $i$. $\mathcal{N}_{i}$ is the neighborhood of robot $i$. For any transmitting robot $j\in\mathcal{N}_i$ we assume that robots $i$ and $j$ can communicate by broadcasting lightweight packets (50 bytes/packet).
%=========System Architecture=======
\begin{figure*}
  \begin{minipage}{.33\linewidth}
    \includegraphics[width=6.0cm,height=5.5cm]{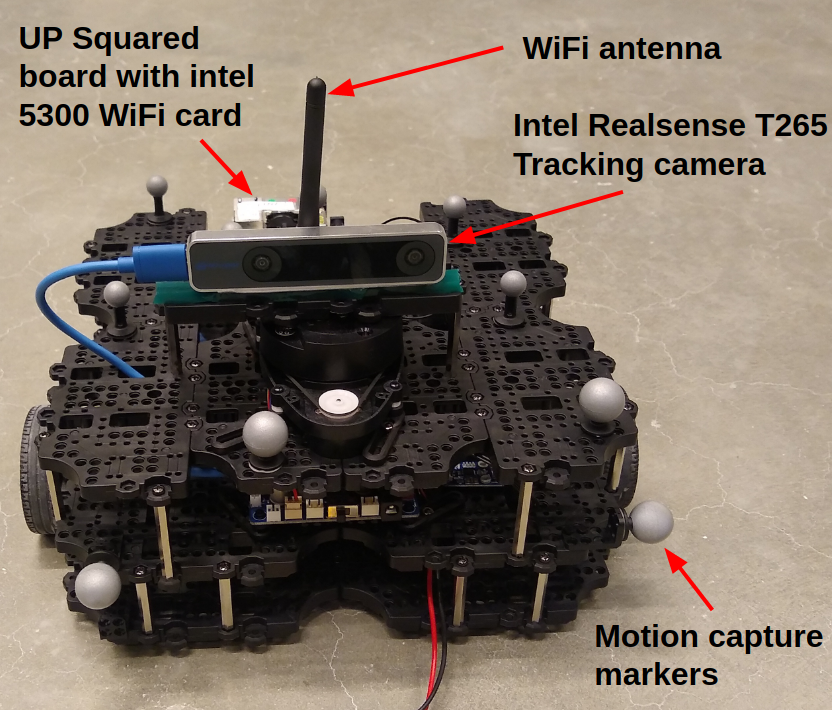}
  \end{minipage}
  \begin{minipage}{.65\linewidth}
%   \vspace{-0.03in}
    \includegraphics[width=11.8cm,height=6.0cm]{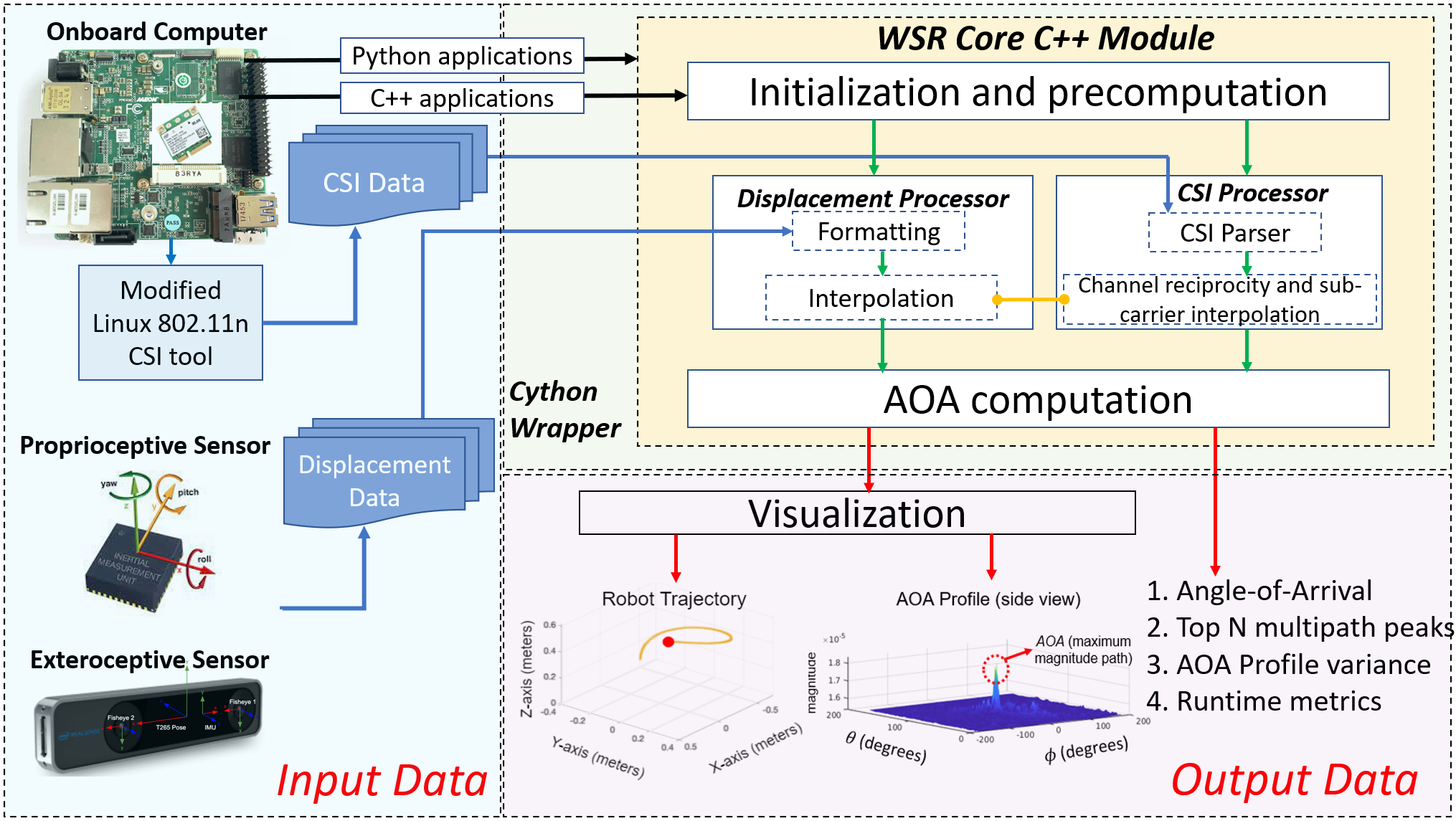}
  \end{minipage}
  \caption{\footnotesize{\emph{\textbf{Left}: Robot platform and sensors used for hardware experiments. \textbf{Right}: WSR toolbox System Architecture. The input data steams, robot displacement and WiFi CSI data are obtained using inertial sensors (exteroceptive and proprioceptive sensors) and commercial WiFi cards (e.g Intel 5300). The core C++ library pre-processes them before computing the AOA profile (Eqn. \ref{eqn:bartlett_estimator}), $AOA_{max}$, Top N peaks, profile variance (Eqn.~\ref{eqn:profile_var}) and runtime metrics.}}}
  \label{fig:sys_arch}
  \vspace{-0.25in}
\end{figure*}
%=========System Architecture=======

Robot $i$ processes $M$ WiFi packets received from robot $j$ as it moves from time $t_k$ to $t_l$ (Fig. \ref{fig:coordinate_frame}). This is akin to simultaneously capturing these packets by a $M$-element \emph{virtual} antenna array. The phase difference of the received signal at this \emph{virtual} antenna array is calculated using the \emph{steering vector} $a(\theta,\phi,d(t))$ that determines array geometry~\cite{Vu2010PerformanceAO}. Angles ($\phi$ (azimuth), $\theta$(elevation)) refer to all possible pairs of candidate directions of robot $j$. $d(t)$ is the robot $i$'s displacement in $\mathbb{R}^3$ from time $t_k$ to any time $t\in[t_k, t_l]$. We then apply a direction of arrival algorithm, the Bartlett estimator~\cite{Krim1996TwoDO}, to obtain the AOA profile ${F_{ij}(\phi,\theta)}$ (Fig. \ref{fig:real_aoa_profile}). It captures the received signal multipaths at robot $i$ from robot $j$ along all directions and is given by:
\vspace{-0.1in}
\begin{equation}
    \label{eqn:bartlett_estimator}
    F_{ij}(\phi,\theta)=%{\left|\sum_{t} \textbf{h_{ij}(t)}*\textbf{a}(\theta,\phi)\textbf{(t)}\right|^2}
    {\left|\sum_{t=t_k}^{t_l} {h_{ij}(t)}~a(\theta,\phi,d(t)) \right|^2} .
\end{equation}

\vspace{-0.05in}
\noindent $h_{ij}(t)$ is the WiFi channel which can be obtained using the signal's Channel State Information (CSI) \cite{Halperin_csitool,nexmon-csitoolbox}. It contains the signal's strength and its phase. We note that the signal power has no impact on ${F_{ij}(\phi,\theta)}$ and acts as only a scaling factor. In the rest of the paper we refer to the wireless signal's information as CSI data. $AOA_{max}$ in ${F_{ij}(\phi,\theta)}$ refers to the strongest signal multipath (often the direct line path) and is used as relative AOA between robots.

%% file: text/4_system_architecture.tex
% \vspace{-0.02in}
\section{System Architecture} \label{sec:sys_arch}
% \vspace{-0.05in}
In this section, we first introduce the major components of the WSR toolbox. Then we focus on the design challenges that our toolbox resolves for distributed and realtime deployment on robot platforms. The toolbox architecture in Fig. \ref{fig:sys_arch} shows three high level components: a) \emph{Input data streams:} this includes robots' displacement and CSI data measured using robots' on-body sensors, b) \emph{Core Module:} processes input data streams using Eq.~\ref{eqn:bartlett_estimator} to calculate AOA profile and c) \emph{Output data:} includes AOA profile ${F_{ij}(\phi,\theta)}$, AOA and performance metrics.
% \begin{itemize}
%     \vspace{-0.04in}
%     \item \emph{Input data streams:} this includes robots' displacement and CSI data measured using robots' on-body sensors.
%     \item \emph{Core Module:} processes input data streams using Eq.~\ref{eqn:bartlett_estimator} to calculate AOA profile. It can be used natively in C++ or with a Cython wrapper. 
%     \item \emph{Output data:} includes AOA profile ${F_{ij}(\phi,\theta)}$, AOA and performance metrics.
%     \vspace{-0.04in}
% \end{itemize}
Detailed technical specifications of these components and other functionalities like debugging and visualization are available on the toolbox's github page. 

Next, we focus on the main design challenges associated with these components and addressed within the toolbox release. These include using a robots' local sensors (Sec.~\ref{sec:sys_local_sensors}), compatibility with multi-robot systems (Sec.~\ref{sec:sys_multirobot_sys}) and leveraging signal multipath (Sec.~\ref{sec:sys_signal_multipath}).

% \vspace{-0.06in}
\subsection{\textbf{Implementation with robots' local sensors}}\label{sec:sys_local_sensors}
% \vspace{-0.04in}
The toolbox input streams are impacted by noise on account of using inertial sensors and off-the-shelf WiFi cards.

\noindent \emph{\underline{Using robots' noisy displacement estimate:}} A robots' on-board inertial sensors are impacted by noise and accumulating drift. However, the robot displacements in $\mathbb{R}^3$ used by our toolbox are of the order of 1m with the minimum being two times the signal wavelength ($\approx$ 12 cm for 5Ghz WiFi)~\cite{orfanidis2016electromagnetic}. In~\cite{WSR_IJRR} we show that a) only the estimation error in displacement between time $t_k$ to $t_l$ impacts AOA and, b) accurate AOA is best achieved with robot motion along a curved path in $\mathbb{R}^3$, since such paths are more \emph{informative} than straight line paths. The AOA accuracy results using local sensors like a tracking camera are discussed in Sec.~\ref{sec:aoa_accuracy_benchmark}.   

\noindent \emph{\underline{Using off-the-shelf WiFi cards}}: The CSI data for commercial WiFi cards is also impacted by noise i.e Carrier Frequency Offset (CFO). This can be corrected by pairing CSI data of WiFi packets broadcasted from robot $i$ and $j\in\mathcal{N}_i$ almost simultaneously. Previous work requires packets to be paired using timestamps (requiring micro-second level synchronization)~\cite{Gil2015AdaptiveCI}. Instead, our toolbox embeds an ID number for each broadcasted packet. The \emph{CSI processor} in Core Module uses this information for pairing the packets. We thus relax the requirement of time synchronization which is challenging to maintain for distributed robot teams. We note that all robots in the team need to install the toolbox on their on-board computers to collect CSI data.

% \vspace{-0.06in}
\subsection{\textbf{Compatibility with multi-robot systems}}\label{sec:sys_multirobot_sys}
\vspace{-0.05in}
This section discusses the challenges pertaining to efficient usage of computing resources available to a robot, performance tradeoffs and scalability. 

\noindent \emph{\underline{Onboard computation and performance tradeoffs:}} Configuring different parameters during \emph{Initialization and precomputation} in Core Module impacts \emph{AOA computation}, modifying the toolbox performance. Computation can be selected to be parallelized for faster processing. The overall parallelization leads to 40\% improvement in runtime as compared to a single-threaded implementation. A part of the steering vector $a(\theta,\phi,d(t))$ calculation is also precomputed leading to a maximum improvement of 10\% in processing time during computation of AOA in a continuous loop. The output AOA profile ${F_{ij}(\phi,\theta)}$ has a default resolution of 360x180 with a granularity of 1 degree i.e 360 degrees in azimuth and 180 degrees in elevation. However this can be modified to improve runtime and memory usage. For example, reducing the resolution of ${F_{ij}(\phi,\theta)}$ can improve processing time by up to 75\% (Table \ref{table:runtime}) with modest increase in AOA error (Fig. \ref{fig:different_config_perf}). Additional details are provided in section \ref{sec:aoa_accuracy_benchmark}.

\noindent \emph{\underline{Scaling to distributed multi-robot systems}} Robots in a team need to simultaneously broadcast WiFi packets when using our toolbox. However this results in indeterministic packet delays and failure to correct CFO. Therefore, we implement a simple round-robin protocol similar to Time-division multiple access (TDMA) algorithm by modifying part of the Linux 802.11n CSI Tool~\cite{Halperin_csitool}. Thus, a robot $j\in\mathcal{N}_i$ transmits packets only when it detects a packet intended for itself that is broadcasted from robot $i$. This simplifies scaling-up the toolbox deployment for multiple robots.

% \vspace{-0.06in}
\subsection{\textbf{Leveraging signal multipath}}\label{sec:sys_signal_multipath}
\vspace{-0.04in}
In this section we discuss how the toolbox can leverage the whole AOA profile ${F_{ij}(\phi,\theta)}$ for rejecting possible outliers to improve AOA estimation robustness. The toolbox is also designed to identify prominent signal multipaths with implications to improving bearing estimation accuracy in NLOS as well as enabling other applications. 

\noindent \emph{\underline{Profile variance:}} The AOA profile can experience multiple peaks due to noisy inputs or signal multipath. Thus, it is important to understand which AOA estimates are outliers and subject to rejection. Thus to determine the reliability of estimated AOA, the \emph{AOA Computation} sub-module returns a \emph{variance} metric $\sigma_{ij}$ for every AOA profile that is calculated. In Sec.~\ref{sec:localization} we show how $\sigma_{ij}$ is used for outlier rejection in a localization task. $\sigma_{ij}$ is the variance of ${F_{ij}(\phi,\theta)}$ around $AOA_{max}$. Extending the metric from \cite{Gil2015AdaptiveCI} to the general 3D case, it is given as:
%===============original code block=================
% \begin{align}\label{eqn:profile_var}
%   F &= \sum_{\phi \in {[-\pi,\pi]}}\sum_{\theta \in {[0,\pi]}}f_{ij}(\phi,\theta) \nonumber \\
%   \sigma_{F_{ij}} &= \sum_{\phi \in {[-\pi,\pi]}}\sum_{\theta \in {[0,\pi]}}\frac{\Psi f_{ij}(\phi,\theta)}{F} \nonumber \\
%   \sigma_{N_{ij}} &= \sum_{\phi \in {[-\pi,\pi]}}\sum_{\theta \in {[0,\pi]}}\frac{\Psi F}{A} \nonumber \\
%   \sigma_{ij} &= \frac{\sigma_{F_{ij}}}{\sigma_{N_{ij}}} ,
% \end{align}
% \noindent where \emph{A} being the number of all possible combinations of $\phi$, $\theta$ and $\Psi=(\phi-\phi_{max})^2+(\theta-\theta_{max})^2$, $\phi_{max}$ and $\theta_{max}$ being the azimuth and elevation angles of $AOA_{max}$.
%===============original code block=================
%===============inline code block=================
\vspace{-0.05in}
\begin{align}\label{eqn:profile_var}
  \sigma_{ij} &= \frac{\sigma_{F_{ij}}}{\sigma_{N_{ij}}} ,
\end{align}

\begin{figure}
    \vspace{0.05in}
    % \centering
    \includegraphics[width=8.5cm,height=5.0cm]{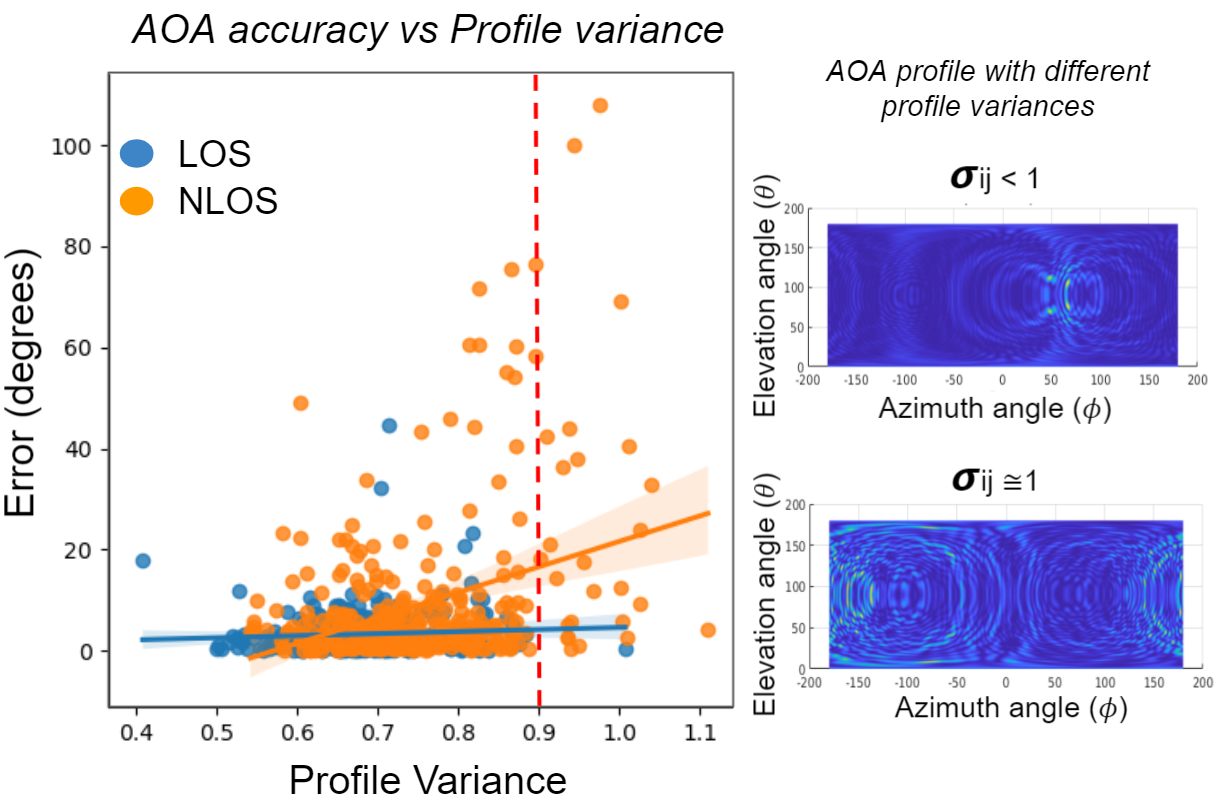} % second figure itself
    \vspace{-0.15in}
    \caption{\footnotesize{\emph{Profile variance $\sigma_{ij}$ (Eq. \ref{eqn:profile_var}) and corresponding AOA error for LOS and NLOS. High AOA error is observed specifically for NLOS samples beyond a threshold of $\sigma_{ij}> $0.9 which indicates a noisy AOA profile. Two sample profiles from hardware experiments, one with a low variance and other with high variance are shown on the right.}}}
    \label{fig:profile_var}
    \vspace{-0.3in}
\end{figure}

\vspace{-0.05in}
\noindent where $\sigma_{F_{ij}} = \sum_{\phi \in {[-\pi,\pi]}}\sum_{\theta \in {[0,\pi]}}\frac{\Psi f_{ij}(\phi,\theta)}{F}$ denotes the sum of how far each peak in ${F_{ij}(\phi,\theta)}$ is away from $AOA_{max}$,  $\sigma_{N_{ij}} = \sum_{\phi \in {[-\pi,\pi]}}\sum_{\theta \in {[0,\pi]}}\frac{\Psi F}{A}$ is the normalization factor with \emph{A} being the number of all possible combinations of $(\phi, \theta)$. $F= \sum_{\phi \in {[-\pi,\pi]}}\sum_{\theta \in {[0,\pi]}}f_{ij}(\phi,\theta)$, where $f_{ij}(\phi,\theta)$ is the magnitude of a peak along a specific direction $(\phi,\theta)$. $\Psi=(\phi-\phi_{max})^2+(\theta-\theta_{max})^2$, $\phi_{max}$ and $\theta_{max}$ being the azimuth and elevation angles of $AOA_{max}$.
%===============inline code block=================
% \noindent where $\sigma_{F_{ij}} = \sum_{\phi \in {[-\pi,\pi]}}\sum_{\theta \in {[0,\pi]}}\frac{\Psi f_{ij}(\phi,\theta)}{F}$, $\sigma_{N_{ij}} = \sum_{\phi \in {[-\pi,\pi]}}\sum_{\theta \in {[0,\pi]}}\frac{\Psi F}{A}$. $f_{ij}(\phi,\theta)$ is the magnitude of a peak along a specific direction $(\phi,\theta)$, $F= \sum_{\phi \in {[-\pi,\pi]}}\sum_{\theta \in {[0,\pi]}}f_{ij}(\phi,\theta)$, \emph{A} being the number of all possible combinations of $\phi$, $\theta$ and $\Psi=(\phi-\phi_{max})^2+(\theta-\theta_{max})^2$, $\phi_{max}$ and $\theta_{max}$ being the azimuth and elevation angles of $AOA_{max}$. 
$\sigma_{ij}$ $<$ 1 corresponds to ${F_{ij}(\phi,\theta)}$ that has very few or no multipaths, $\sigma_{ij}$ $\approx$ 1 corresponds to a noisy ${F_{ij}(\phi,\theta)}$ and $\sigma_{ij}$ $>$ 1 implies the presence of strong multipaths. This allows us to reject estimates of $AOA_{max}$ that are more likely to be erroneous e.g when $\sigma_{ij}\approx$ 1. The toolbox allows setting a user-defined variance threshold value $\tau$ to reject bearing estimates for $\sigma_{ij}>\tau$. An example of this is show in Fig.~\ref{fig:profile_var} for data collected from our hardware experiments. A profile with lower variance is less noisy as compared to the one with high variance and hence is more reliable for AOA estimation.

\noindent \emph{\underline{Direct signal path versus multipath:}} $AOA_{max}$ corresponds to the dominant signal direction in ${F_{ij}(\phi,\theta)}$ and is used as relative bearing between robots. To enable additional applications, the toolbox also returns other prominent signal multipaths when $\sigma_{ij}<$1 (Eq.~\ref{eqn:profile_var}). In practice, true multipath peaks in ${F_{ij}(\phi,\theta)}$ are closer in magnitude compared to peaks due to noise (Fig.~\ref{fig:real_aoa_profile})~\cite{music,root-music}. Thus, the $Top~N$ peaks in ${F_{ij}(\phi,\theta)}$ are selected such that they are at least $K$\% of the $AOA_{max}$'s magnitude. Each of $Top~N$ peaks are also required to be distinct multipaths and not local maxima around an existing peak, like $AOA_{max}$. This is achieved by ensuring that no two peaks in $Top~N$ are within $\alpha$ degrees of each other. $N$, $K$ and $\alpha$ are chosen empirically depending on the extent of multipath richness of a given environment (i.e. how cluttered it is). Potential uses of the $Top~N$ peaks for different multi-robot applications is discussed in Sec.~\ref{sec:enablings_appplications}.

%====================Figure - Profile variance ==================
%====================Figure - Profile variance ==================
%======================= Testbed setup===================
\begin{figure*}[h]
\centering
  \begin{minipage}{.65\linewidth}
  \includegraphics[width=11.6cm,height=5.5cm]{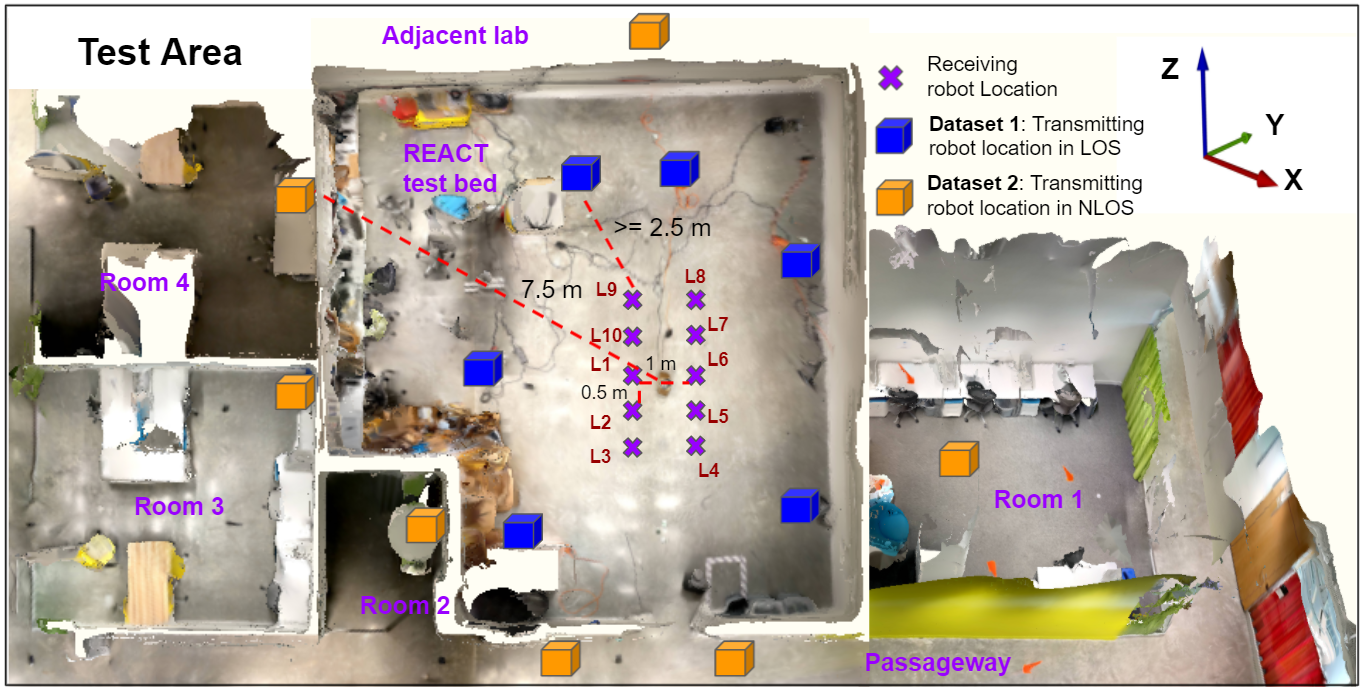}
  \end{minipage}
  \begin{minipage}{.33\linewidth}
  \vspace{0.15in}
  \includegraphics[width=5.75cm,height=5.0cm]{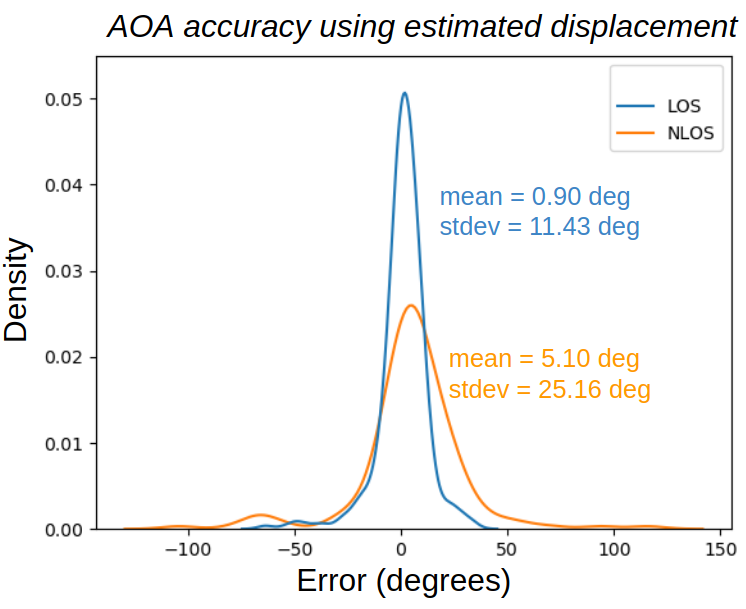}
  \end{minipage}
  \caption{\footnotesize{\emph{\textbf{Left}: Test area used for experiments. \textbf{Right}: KDE plot for AOA error in LOS and NLOS scenarios, using estimated displacement and applying outlier rejection based on profile variance (Eqn. \ref{eqn:profile_var}). From these empirical results, we can approximate the AOA error of our system as a Gaussian distribution.}}}
  \label{fig:test_area_map}
  \vspace{-0.25in}
\end{figure*}
%======================= Testbed setup===================

%% file: text/5_results.tex
% \vspace{-0.05in}
\section{Experimental Evaluation}
% \vspace{-0.05in}
This section gives details of the robot platforms and sensors used for the hardware experiments and results of AOA accuracy. We place the robots in LOS and NLOS of each other in different rooms of an indoor office building. We also provide performance results for onboard computation on the Up-Squared board. To showcase the toolbox utility we conduct a multi-robot localization experiment and evaluate the localization error resulting from estimated AOA. Furthermore, we also release the dataset of these experiments to motivate further research within the community.

% \vspace{-0.1in}
\subsection{\textbf{Hardware and Testbed Setup}}\label{sec:exp_setup}
\vspace{-0.05in}
We use Turtlebot3 ground robots equipped with UP-Squared board (8GB RAM) and a 2dBi WiFi antenna. The signal phase is obtained from CSI data collected using Intel 5300 WiFi card and customized Linux 802.11n CSI Tool. Localizing robot $i$'s displacements are captured using motion capture system (baseline), on-board odometer and Intel T265 tracking camera (estimated displacement). The testbed setup is shown in Fig.~\ref{fig:test_area_map}. The total test area is approx. 155 sq. meters and spans 7 locations in the office area. These include the main testbed with motion capture system and 6 locations that are in NLOS. The locations include chairs, tables, glass door, electronics and metal shelves among others. We collect data samples for ten positions of robot $i$ arranged in a grid. These positions are at a minimum distance of 2.5m from LOS robots in $\mathcal{N}_i$. For NLOS, robots in $\mathcal{N}_i$ are placed in adjacent office spaces at a maximum distance of $\approx$ 8 m. The true positions of robots in $\mathcal{N}_i$ are known and are calculated using motion capture system, measuring tape, laser range pointer and 3D RGBD map generated using an iphone.
%====================Figure - performance evaluation ==================
\begin{figure}
    \centering
    \includegraphics[width=6.5cm,height=5.0cm]{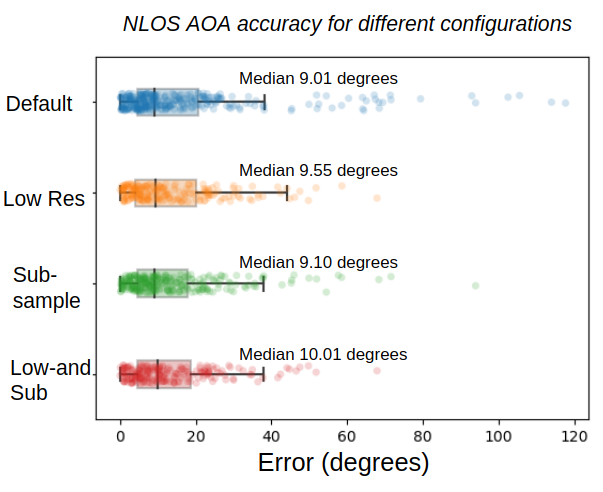} % second figure itself
    \vspace{-0.15in}
    \caption{\footnotesize{
    \emph{AOA accuracy in NLOS using estimated displacement for different computation configurations (Table \ref{table:runtime}). Although there is significant difference in runtime, their AOA estimation errors are comparable.}}}
    \label{fig:different_config_perf}
    \vspace{-0.05in}
\end{figure}
%====================Figure - performance evaluation ==================
%====================Table - performance evaluation ==================
\begin{table}
\caption{Runtime comparison using multi-threaded implementation.}
\vspace{-0.2in}
\label{table:runtime}
\begin{center}
\begin{tabular}{| c | c | c | c | c |}
\hline
\textbf{Configuration} & \multicolumn{2}{ c |}{\textbf{Parameters}} & \multicolumn{2}{ c |}{\textbf{Runtime time (sec)}}  \\ 
\cline{2-5}
& \textbf{Avg Pkt} & \textbf{Res} & \textbf{Laptop} & \textbf{UP board}\\
\hline
Default & 880 & 360x180 & 5.3 & N/A\\ \hline
Low-Res & 880 & 180x90 & 1.33 & 3  \\ \hline
Sub-sample & 440 & 360x180 & 2.67 & 7  \\ \hline
Low-and-Sub & 440 & 180x90 & 0.67 & 1.75 \\ \hline
\end{tabular}
\end{center}
\vspace{-0.35in}
\end{table}
%====================Table - performance evaluation ==================
% \vspace{-0.09in}
\subsection{\textbf{Bearing (AOA) benchmark}}\label{sec:aoa_accuracy_benchmark}
\vspace{-0.05in}
\noindent \emph{\underline{Accuracy evaluation:}}
We evaluate the toolbox's performance using a ground robot. When an AOA measurement is required, the robot $i$ broadcasts packets that simultaneously activates the auto-response of robots in $\mathcal{N}_i$. The displacement of robot $i$ and CSI data for robots $i$ and $j\in\mathcal{N}_i$ are collected on-board in realtime. AOA accuracy is measured by the difference between $AOA_{max}$ and groundtruth AOA in x-y plane between a pair robots $i$ and $j$. A total of 637 samples are collected at different setup locations (Sec.~\ref{sec:exp_setup}). We do not reject any samples during data collection. The rejection threshold is based on profile variance $\sigma_{ij}$ (Eq.~\ref{eqn:profile_var}). It is determined empirically and applied uniformly to all collected data samples before computing aggregate AOA results. For 294 LOS and 343 NLOS samples, any sample with $\sigma_{ij}>$ 0.9 is identified as an outliers. Thus, 0.34\% of LOS samples and 7.5\% of NLOS samples are rejected. The mean error of these rejected samples is 65.31 degrees, indicating good correlation between AOA accuracy and $\sigma_{ij}$ (Fig.~\ref{fig:profile_var}). Figure~\ref{fig:test_area_map} shows the KDE plot of AOA accuracy, using estimated robot displacement from the tracking camera. As indicated from these results, the error in LOS and NLOS can be approximated by a Gaussian distribution. Additional AOA accuracy results are available at the toolbox's github repository page.  

\noindent \emph{\underline{Performance tradeoff:}} We also compare the runtime performance of the toolbox on the UP-Squared board to that of a laptop (with 8 cores and 64 GB RAM) for different computation configurations (Table \ref{table:runtime}). These include:
\begin{itemize}
    \item \emph{Default:} AOA profile resolution 360x180, uses all collected packets.
    \item \emph{Low Res:} resolution 180x90, uses all packets.
    \item  \emph{Sub-sample:} resolution 360x180, sub-sample packets (i.e. using every alternate packet).
    \item \emph{Low-and-Sub:} resolution 180x90, sub-sample packets.
\end{itemize}
% a) \emph{Default:} AOA profile resolution 360x180, uses all collected packets, b) \emph{Low Res:} resolution 180x90, uses all collected packets, c) \emph{Sub-sample:} resolution 360x180, sub-sample the packets (using every alternate packet) and d) \emph{Low-and-Sub:} resolution 180x90, sub-sample packets. 
For \emph{Default} configuration, the maximum packets process on the Up-Squared board are limited to $\approx$ 450 packets due to memory constraints. An improvement in processing time and max allowable packets is observed for configuration \emph{Low-and-Sub}. Using configurations other than the default may lead to increased profile variance due to computation approximations, resulting in higher sample rejection.
% \vspace{-0.1in}
\subsection{\textbf{Multi-robot localization}}\label{sec:localization}
\vspace{-0.05in}
We showcase the utility of our system for a multi-robot localization scenario. A ground robot $i$ needs to localize itself with respect to transmitting robots in $\mathcal{N}_i$ whose positions are known but can be in NLOS to the localizing robot $i$ (Fig.~\ref{fig:test_area_map}). We evaluate two configurations in both LOS and NLOS: a)~\emph{convex-hull}: using all 7 robots in $\mathcal{N}_i$; this arrangement of robots provides coverage over the convex hull for localizing robot $i$. b)~\emph{non-convex-hull}: using 6 robots in $\mathcal{N}_i$. We first compute the AOA of all robots in $\mathcal{N}_i$ relative to robot $i$, online and then run the localization algorithm.

\begin{figure}
    \begin{minipage}{0.47\textwidth}
        \centering
        \includegraphics[width=9.0cm,height=5.7cm]{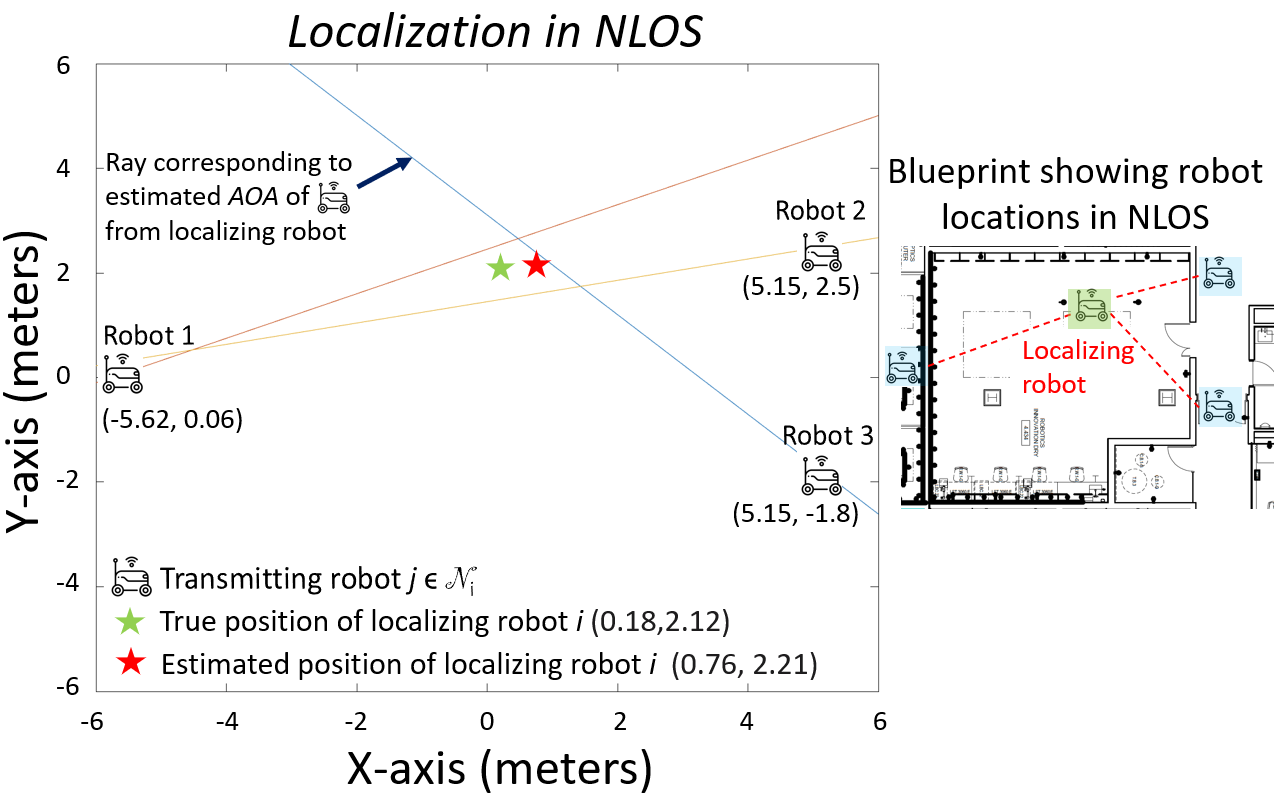} % second figure itself
    \end{minipage}
    \caption{\footnotesize{\emph{Shows an example using the least-square intersection method to localize a receiving robot $i$ using estimated AOA to transmitting robots $j\in\mathcal{N}_i$. The blueprint on the right shows the corresponding locations of the robots that in different rooms and all in NLOS to one another.}}}
    \label{fig:localization_algorithm}
    \vspace{-0.06in}
\end{figure}

\noindent \emph{\underline{Localization Algorithm:}} 
The localization algorithm runs offline and uses these AOA estimates to estimate the position of robot $i$. We use a least-square intersection algorithm. It takes as input the position of robots in $\mathcal{N}_i$ and their AOA estimated by the robot $i$. This allows for generating a \emph{ray} that originates from robot $j\in\mathcal{N}_i$ to robot $i$. Then the point that minimizes the sum of squared distances to all the rays is treated as the predicted position of the robot $i$ (Fig.~\ref{fig:localization_algorithm}). To be more specific, the line across a point $\textbf{a}_j$ for $j\in\mathcal{N}_i$ and along a unit vector $\textbf{n}_j$ formed by $AOA_{max}$ can be represented as $\textbf{a}_j + \lambda\textbf{n}_j$. Then we can write $\textbf{n}_j$ in the form of $AOA_{max}$: $\textbf{n}_j = [\cos(AOA_{max}), \sin(AOA_{max})]^T$. The distance from the localizing robot $i$'s position $\textbf{p}_i$ to the line can be written as vector form: $D_{ij}(\textbf{p}_i) = ||(\textbf{p}_i-\textbf{a}_j)-((\textbf{p}_i-\textbf{a}_j) \cdot \textbf{n}_j)\textbf{n}_j||$. The localization problem can thus be represented as a least square problem: $\underset{\textbf{p}_i}{\arg\min} \sum_{j \in \mathcal{N}_i}D_{ij}^2(\textbf{p}_i)$. We solve this least square problem to localize robot $i$. (Fig.~\ref{fig:localization_algorithm}). 
%====================Fig-Localization Algorithm ==================
%====================Fig-Localization Algorithm ==================
%====================Rendezvous example ==================
\begin{figure}
    \centering
    \begin{minipage}{0.47\textwidth}
        \centering
        \includegraphics[width=7.5cm,height=3.5cm]{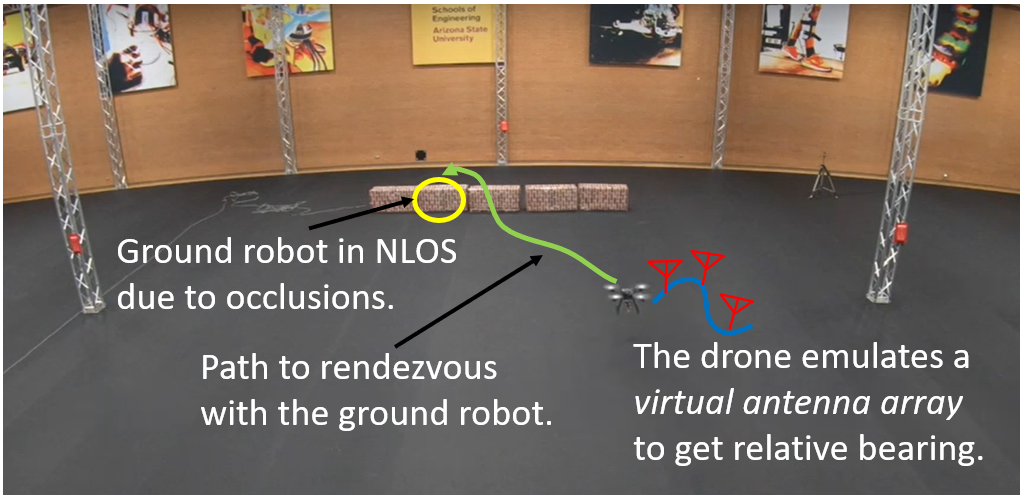} % second figure itself
    \end{minipage}
    \caption{\footnotesize{\emph{Dynamic rendezvous between a UAV and ground robot using AOA from our system. Video link: \href{https://git.io/JuKOS}{https://git.io/JuKOS}}}}
    \label{fig:rendezvous}
    \vspace{-0.2in}
\end{figure}
%====================Rendezvous example ==================

\begin{figure}
    \centering
    \begin{minipage}{0.47\textwidth}
        \centering
        \includegraphics[width=7.0cm,height=5.5cm]{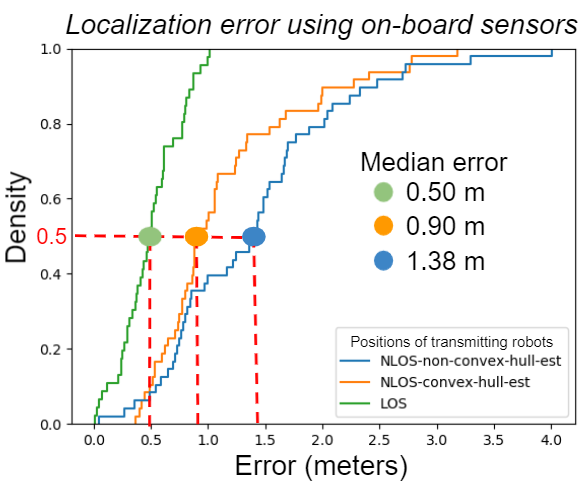} % second figure itself
    \end{minipage}
    \caption{\footnotesize{\emph{Localization accuracy for a ground robot in LOS and NLOS using AOA from all on-board sensors. We reject samples using the profile variance $\sigma_{ij}$ of 0.9 (Eq. \ref{eqn:profile_var}) to achieve sub-meter median error for LOS and NLOS convex-hull case. (See Fig. \ref{fig:test_area_map} and Sec.~\ref{sec:localization} for setup details.)}}}
    \label{fig:localization}
    \vspace{-0.25in}
\end{figure}

\noindent \emph{\underline{Evaluation:}} Potential outlying AOA measurements are rejected using profile variance $\sigma_{ij} =$ 0.9 (See Eq.~\ref{eqn:profile_var}). Aggregate results for localization accuracy in NLOS using estimated displacement of robot $i$ are shown in Figure~\ref{fig:localization}. The \emph{convex-hull} setup achieves 0.9 m median accuracy. In comparison, the median accuracy in LOS is 0.5 m. The \emph{non-convex-hull} setup for NLOS leads to $\approx$ 0.45m higher median localization error compared to that of convex-hull setup.
% \vspace{-0.075in}
\subsection{\textbf{Other applications}}\label{sec:enablings_appplications}
\vspace{-0.05in}
In this section we discuss other possible uses of the full AOA profile, particularly the multipath peaks, provided by our WSR toolbox. For localization use cases, having the full AOA profile could allow for better estimation of the relative bearing between robots. For example, the $Top~N$ peaks in ${F_{ij}(\phi,\theta)}$ could help in identifying the direct signal path more accurately. One idea is to extract the direct-path from ${F_{ij}(\phi,\theta)}$ by observing those peaks in $Top~N$ that undergo minimal change proportional to the robots' motion which is characteristic of the direct path, in contrast to multipath which change in a non-deterministic way~\cite{Goldsmith2005WirelessC}. Other applications that exploit the measuring of all signal paths include rendezvous (Fig.~\ref{fig:rendezvous}), ad-hoc mobile networks~\cite{Gil2015AdaptiveCI,robot-networks}, and security and resilience in robot teams where the full AOA profile is used as a spatial \emph{fingerprint}~\cite{Gil2015Spoof-Resilient,crowd_vetting}.  

% \vspace{-0.05in}
\section{Datasets}
% \vspace{-0.05in}
We release the data of our hardware experiments to enable offline testing, simulation of an NLOS wireless bearing sensor, or potentially machine learning applications. It includes two subsets - LOS and NLOS. Each subset includes ten locations of the receiving robot $i$ along a grid given fixed known locations of three transmitting robots $j\in\mathcal{N}_i$. The collected data consists of CSI data, groundtruth displacements using motion capture system and estimated displacements using a T265 tracking camera as well as wheel odometry of a Turtlebot3 robot. The CSI data consists of $\approx$880 packets per AOA profile. The data is collected as follows: 
\begin{itemize}
    \vspace{-0.04in}
    \item Packet transmission rate: 200 packets/sec (4 sec)
    \item 2D robot displacement: linear velocity = 0.2 m/sec, angular velocity = 0.4 m/sec
    \vspace{-0.04in}
\end{itemize}
\noindent We also include the calculated AOA and performance metrics for these datasets which are provided in json format.

%====================Fig-Localization Accuracy CDF ==================
%====================Fig-Localization Accuracy CDF ==================

%% file: text/7_conclusion.tex
% \vspace{-0.05in}
\section{Conclusion}
% \vspace{-0.05in}
This paper presents the WSR toolbox for computing relative bearing between robots in a multi-robot setting using their displacements and WiFi signal data. We release pertinent datasets for many scenarios of operation including LOS and NLOS environments and a localization usecase for which we include performance results. We hope that this toolbox provides the robotics community with new perception capabilities using WiFi-as-a-Sensor in general, NLOS and GPS-denied environments with implications for localization, adhoc robot networks, and security of multi-robot teams amongst other potential uses. 

\noindent \textbf{Acknowledgements:} We are thankful to Todd Zickler for providing access to his lab during NLOS experiments.